\documentclass[runningheads]{llncs}
\usepackage{graphicx}
\usepackage{amssymb}
\usepackage{amsmath} % define this before the line numbering.
\usepackage{color}
\usepackage{color}
\usepackage{verbatim}
\usepackage[width=122mm,left=12mm,paperwidth=146mm,height=193mm,top=12mm,paperheight=217mm]{geometry}

\usepackage{array}
\usepackage{mathrsfs}
\usepackage{multirow}

\begin{document}
% \linenumbers
\pagestyle{headings}
\mainmatter

\title{A Multi-Stage Multi-Task Neural Network for Aerial Scene Interpretation and Geolocalization} % Replace with your title

%\titlerunning{A very long title}

%\authorrunning{authors running}

\author{Authors}

%Please write out author names in full in the paper, i.e. full given and family names. 
%If any authors have names that can be parsed into FirstName LastName in multiple ways, please include the correct parsing, in a comment to the volume editors:
%\index{Lastnames, Firstnames}
%(Do not uncomment it, because you may introduce extra index items if you do that...)

\author{Alina Marcu\textsuperscript{1,3} \and Dragos Costea\textsuperscript{2,3}
\and Emil Slusanschi\textsuperscript{2} \and
Marius Leordeanu\textsuperscript{1,2,3}
}

\institute{\textsuperscript{1} "Simion Stoilow" Institute of Mathematics of the Romanian Academy \\
\textsuperscript{2} University Politehnica of Bucharest, Romania\\
\textsuperscript{3} Autonomous Systems, Bucharest, Romania\\
	\email{ \{alina.marcu, dragos.costea\}@autonomous.ro \\ \{emil.slusanschi, marius.leordeanu\}@cs.pub.ro}
}

\maketitle

\begin{abstract}

Semantic segmentation and vision-based geolocalization in aerial images are challenging tasks in computer vision.  Due to the advent of deep convolutional nets and the availability of relatively low cost UAVs, they are currently generating a growing attention in the field. We propose a novel multi-task multi-stage neural network that is able to handle the two problems at the same time, in a single forward pass. The first stage of our network predicts pixelwise class labels, while the second stage provides a precise location using two branches. One branch uses a regression network, while the other is used to predict a location map trained as a segmentation task. From a structural point of view, our architecture uses encoder-decoder modules at each stage, having the same encoder structure re-used. Furthermore, its size is limited to be tractable on an embedded GPU. We achieve commercial GPS-level localization accuracy from satellite images with spatial resolution of 1m$^2$ per pixel in a city-wide area of interest. On the task of semantic segmentation, we obtain state-of-the-art results on two challenging datasets, the Inria Aerial Image Labeling dataset and Massachusetts Buildings.

\end{abstract}

\section{Introduction}
\label{sec:intro}

Humans are instantly aware of their surroundings, being able to estimate their approximate position based on a handful of visual cues. With the advent of deep convolutional networks (\cite{he2016deep},\cite{huang2017densely},\cite{yu2015multi}), solutions to such vision problems are now in sight, having the potential to impact technology in the growing fields of autonomous ground or aerial vehicles. Thus, the ability to interpret a scene and tell its location is of growing interest in the domain of aerial images. While the task of semantic segmentation has already attracted many solid approaches, the task of vision-based localization in aerial images is still in its infancy.

In this paper we proposed a novel and competitive approach, a multi-stage, multi-task convolutional neural network, that can solve both semantic segmentation and localization in a single forward pass, using only RGB input. The structure has a singular encoder module, termed UniEncoder, which we prove experimentally to be efficient in learning meaningful descriptors for both pixelwise predictions and localization. Our system is able to produce state-of-the-art segmentation results on several public datasets. The localization precision within a city-wide area is comparable to commercial GPS\cite{Diggelen2015gpsmooc}, making our system suitable for deployment when the airborne sensors malfunction. 

\section{Related Work}
\label{sec:related_work}

Our work is mostly related to approaches in computer vision that address the problems of aerial scene segmentation and geolocalization, as well as work done in designing multi-task networks. For the general case of semantic segmentation, there are many influential papers published. The seminal work on Fully Convolutional Networks(FCNs)~\cite{long2015fully} was among the first to provide dense class labels by using a multi-scale upsampling scheme. This was followed by the U-net model~\cite{ronneberger2015u} with skip connections added, now being one of the most widely used~\cite{garcia2017review}. 
More recent models have refined semantic segmentation into instance segmentation~\cite{he2017mask}, attempting to improve detection performance for overlapping classes.

\vspace{0.15cm}
\noindent\textbf{Semantic segmentation in aerial images.} Inspired by the success of FCNs for semantic segmentation, several authors proposed adaptations of its architecture for aerial images~\cite{sherrah2016fully}. A multi-task learning approach for segmentation in aerial images was introduced in~\cite{bischke2017multi}. While the final aim was the detection of buildings, a distance transform on the buildings labels was computed at an earlier stage. In another approach~\cite{AAAIW1715177} the authors proposed a local-global network with ResNet-like connections~\cite{he2016deep} and two pathways, each with its own depth and structure, in order to treat differently the local information versus the contextual scene.
Network ensembles that generate helpful representations (such as class boundaries) have been proposed, albeit with a prohibitive computational cost and modest qualitative improvements~\cite{marmanis2018classification} over previous state-of-the-art results. 

\vspace{0.15cm}
\noindent\textbf{Visual-based localization.} The general problem of visual-based localization is to determine the position of a query set of data~\cite{piasco2018survey}.
For ground imagery and indoor environments, the simultaneous localization and mapping algorithm is the main contender~\cite{engel2014lsd}. In the special case of monocular camera solutions, feature-based methods achieve state-of-the art accuracy~\cite{mur2017orb}. Some ground, single-image approaches aim to get an accurate location by fitting a box to a building, when an approximate location is known \cite{armagan2017learning}. Binary descriptors have been proposed as a lightweight, human-inspired ground image navigation system~\cite{panphattarasap2018automated}. Planet-wide localization of a photo was also investigated~\cite{weyand2016planet}, the output being a probability function over the surface of the Earth -- with several partitioning schemes being proposed.

In the case of aerial images, the literature is very sparse. Several approaches such as cross-view geolocalization~\cite{lin2013cross} proposed combining aerial with imagery for ground localization. 
Aerial imagery-targeted approaches generally focus on extracting image features \cite{wan2016illumination}. Several high-level features, such as roads and intersections, have been proven to be used successfully in localization~\cite{costea2016aerial,stanfordUAVNavigation2017}. However, they employ complex pipelines and are applied to limited use-cases.

\vspace{0.15cm}
\noindent\textbf{Multi-task networks.} While many multi-task architectures have been proposed for visual recognition (\cite{bansal2017pixelnet}, \cite{kokkinos2016ubernet}), they do not follow a modular stage-wide strategy. A more common approach is to use the same set of learned features and adapt multiple heads, specialized for the different tasks (\cite{collobert2008unified}, \cite{ruder2017overview}). This approach is useful when the output channels of the two tasks are strongly related at the pixel level, such as it is the case with depth, segmentation or surface normals \cite{bansal2017pixelnet}. These tasks benefit from deep common features.

However, in our case, we aim to solve two different problems that do not benefit from direct pixel correspondences. The segmentation network makes its decisions based on pixel-level information and results in a label for the RGB image. The localization network makes use of the whole input scene to generate a global location that is an abstract representation of the input. Therefore, we take a modular approach, with one stage for segmentation and subsequent stages for geolocalization. What is shared among these tasks is the structure of the encoder, not the learned weights.

\section{Our Muti-Stage Multi-Task Architecture}
\label{sec:mtms_architecture}

\begin{figure*}[h!]
%\centering
\includegraphics[scale=0.225]{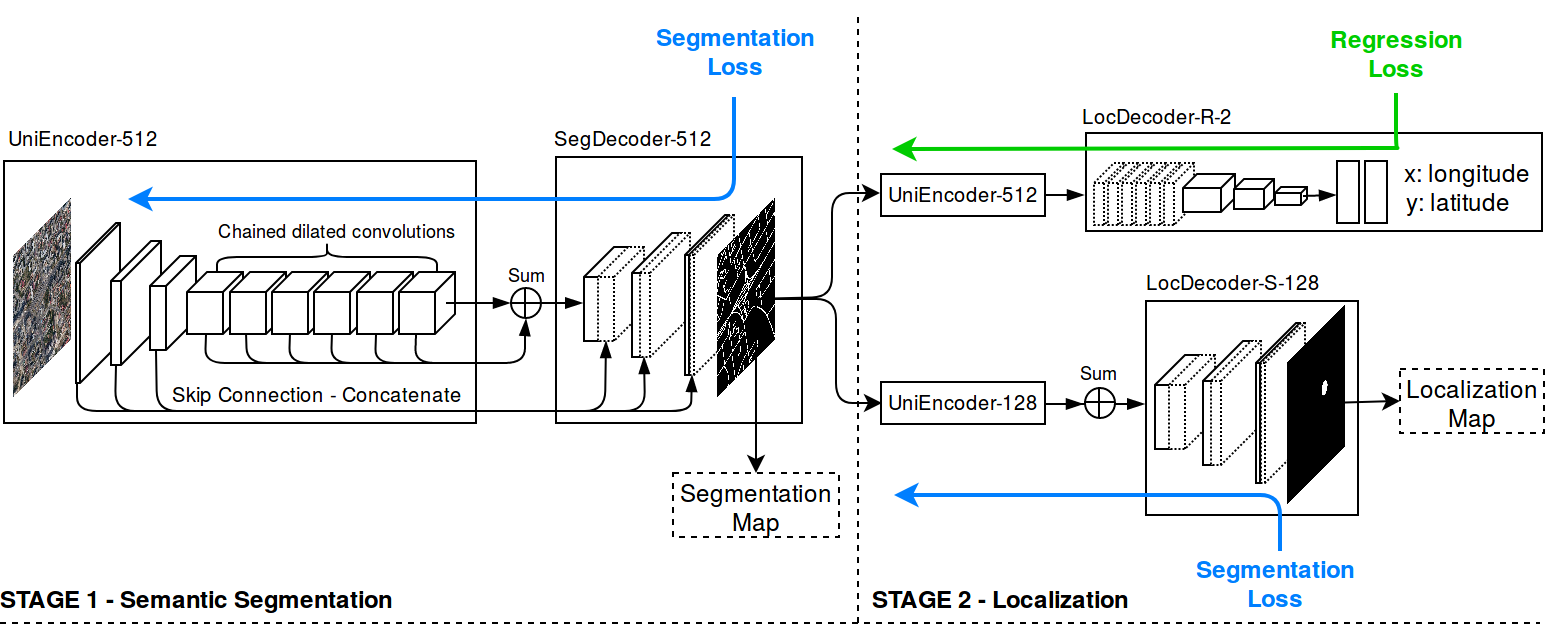}
\caption{\label{fig:msmt_v1} Our proposed multi-stage multi-task (MSMT) architecture for semantic segmentation and geolocalization. While the first stage is designed for semantic segmentation, the second is dedicated to localization, through two branches. For clarity, we put the input size in the name of our encoders and the output size in the name of the decoders. LocDecoder-R-2 predicts location as two real valued numbers for longitude and latitude. LocDecoder-S-128 predicts a localization map of size 128x128 on the whole area of possible locations, where white pixels denote likely locations of the input image. The dotted volumes represent duplicated and concatenated features.
}
\end{figure*}

We propose a multi-stage multi-task (MSMT) architecture, which tackles aerial image segmentation and geolocalization in a single pass. Our approach to solve the two problems within a single network is completely novel in the literature. While the network has several stages and branches, we kept the same encoder and designed the decoders to solve a specific task. The full workflow of our architecture is shown in Figure \ref{fig:msmt_v1}.

The first stage is designed for segmentation -- given an RGB image as input the network produces a semantic map. We train it for road detection -- we argue that roads can be used as a unique footprint of an urban area. Furthermore, it is invariant to environmental conditions, such as illumination and seasonal changes and could be extracted using the same RGB input even for nighttime, provided the training images are adapted to this scenario.

The road segmentation is given as input to the second stage that learns to map it to a location, within a specific city-wide area that it was trained on. For this second stage, we use two branches. The first branch performs localization as regression, where a longitude and latitude pair is outputted for a particular image. The second branch, which is more accurate in experiments, predicts localization as segmentation -- a map of the whole area where white pixels belong to possible locations. The potential drawback of the second branch that performs localization as segmentation, is that there is a possibility of producing multiple potential locations or even more problematic, no location at all, as we will see in the experiments. In these situations, we use the output of the regression network to generate the final result.

\setlength{\tabcolsep}{0.5pt}
\begin{table}
\begin{center}
\caption{Our UniEncoder network structure.}
\label{table:uniencoder_structure}
\centering\begin{tabular}{>{\centering\arraybackslash}m{2cm}>{\centering\arraybackslash}m{2cm}>{\centering\arraybackslash}m{2cm}>{\centering\arraybackslash}m{2cm}>{\centering\arraybackslash}m{2cm}>{\centering\arraybackslash}m{2cm}}
\hline\noalign{\smallskip}
Layer type & Kernel Size & Filters & Dilation rate & Stride & Output Size \\
\noalign{\smallskip}
\hline
\noalign{\smallskip}
Input & & & & & 512x512x3\\
\hline
Conv & 3 x 3 & 64 & 1 & 1 & 512x512x64\\
Conv & 3 x 3 & 64 & 1 & 1 & 512x512x64\\
MaxPool & 2 x 2 &  &  & 2 & 256x256x64\\
\hline
Conv & 3 x 3 & 128 & 1 & 1 & 256x256x128\\
Conv & 3 x 3 & 128 & 1 & 1 & 256x256x128\\
MaxPool & 2 x 2 &  &  & 2 & 128x128x128\\
\hline
Conv & 3 x 3 & 256 & 1 & 1 & 128x128x256\\
Conv & 3 x 3 & 256 & 1 & 1 & 128x128x256\\
MaxPool & 2 x 2 &  &  & 2 & 64x64x256\\
\hline
Conv & 3 x 3 & 512 & 1 & 1 & 64x64x512\\
Conv & 3 x 3 & 512 & 2 & 1 & 64x64x512\\
Conv & 3 x 3 & 512 & 4 & 1 & 64x64x512\\
Conv & 3 x 3 & 512 & 8 & 1 & 64x64x512\\
Conv & 3 x 3 & 512 & 16 & 1 & 64x64x512\\
Conv & 3 x 3 & 512 & 32 & 1 & 64x64x512\\
\hline
\end{tabular}
\end{center}
\vspace*{-10mm}%Put here to reduce too much white space after table 
\end{table}
\setlength{\tabcolsep}{0.5pt}

\subsection{UniEncoder}

The architecture of the encoder used at all stages, i.e. the UniEncoder, is inspired from the work of Yu et al.~\cite{yu2015multi}. The overall MSMT-Net is based, at each stage, on modules of the encoder-decoder net type, frequently used for various segmentation tasks. In the case of the UniEncoder, we reduce our input resolution by a factor of 8 using 2x2 max-pooling layers with stride 2 after two sequential convolutions. For example, for an input of 512x512 the output will be 64x64. We learn filters of size 3x3 and apply ReLU non-linearity after each convolution. Similar to VGG~\cite{simonyan2014very}, after every resolution reduction we double the number of learned filters to reduce information loss. 

Upon reaching our lowest resolution, we process 256 feature maps by aggregating contextual information at progressively increasing scale without losing resolution. We do so by applying 6 chained convolutions exponentially increasing the dilation rate from a dilation of 1 up to 32. The fully-convolutional nature of our segmentation modules permits to have inputs of any size, but we choose to present the details of our architecture for an input size of 512x512. Full technical details of our UniEncoder are given in Table~\ref{table:uniencoder_structure}. 

\subsection{Segmentation Decoder}

After passing the input through the UniEncoder, we feed the segmentation decoder the high-level features produced by the contracting path to generate corresponding pixelwise semantic labels. By inverting the operations of the UniEncoder, we are able to expand low resolution feature maps up to the desired size.
We reuse all the information generated by the chain of dilated convolutions by summing up their output. After this, we proceed with the expansion of the feature maps. We double the resolution in both dimensions by applying 2x2 "up-convolutions", followed by two 3x3 standard convolutions with their corresponding ReLU activations. We make use of the U-Net design \cite{ronneberger2015u} for our downsampling and upsampling paths and concatenate the feature maps from the UniEncoder with the outputs of same size of our segmentation decoder. This is common practice in segmentation networks and has been shown to improve the quality of the segmentations (\cite{kokkinos2016ubernet}, \cite{drozdzal2016importance}). 
We map each feature vector to the desired number of classes by applying a 1x1 convolution as our final layer. 

\subsection{Regression Decoder}

Different from the segmentation decoder, this type of decoder has a fixed output. Also, instead of simply adding all the extracted feature maps from the chain of dilated convolutions, we use the features by concatenating them. We lower the resolution even more with 2-strided convolutions, followed by two 3x3 1-strided convolutions. We apply this block of operations three times. Then we use two fully-connected layers with 512 activation units each, and end the network with a 2-unit fully-connected classification layer. Each layer in the regression decoder is followed by ReLU non-linearities, except for the final classification layer.

\section{Multi-Stage Multi-Task Training}

Stage training is done alternatively. We train our segmentation stage until convergence and then freeze the weights. We use the segmentation maps for the next stage -- localization. Both branches in the second stage depend on the output of the first. Both localization methods are trained independently, with no shared weights.

\subsection{Loss functions}

The performance of deep-learning frameworks relies not only on the choice of network architecture but also on the choice of the loss function. For the task of semantic segmentation, as well as for localization as segmentation, we linearly combine the binary cross-entropy and dice loss functions. We denote $\hat{y_{i}}$ to be the $i^{th}$ output of the last network layer passed through a sigmoid non-linearity, with $\hat{y_{i}} \in [0, 1]$ and $y_{i}$ the corresponding ground truth label with $y_{i} \in \{0, 1\}$. Equation \ref{eq:bce_formula} and \ref{eq:dice_formula} define binary cross-entropy loss and dice loss, respectively.

\begin{equation} \label{eq:bce_formula}
\mathcal{L}_{\textit{bce}} = \sum_{i} y_{i} \log \hat{y_{i}} + (1 - y_{i}) \log (1 - \hat{y_{i}})
\end{equation}

\begin{equation} \label{eq:dice_formula}
\mathcal{L}_{\textit{dice}} = - \frac{2 \sum_{i} \hat{y_{i}}y_{i}}{\sum_{i} \hat{y_{i}} + \sum_{i} y_{i}}
\end{equation}

The learning task of the segmentation pathways (both at the first and second stages) is to minimize the sum of these two losses: 

\begin{equation} \label{eq:sum_formula}
\mathcal{L}_{\textit{segmentation}} = \mathcal{L}_{\textit{bce}} + \mathcal{L}_{\textit{dice}}
\end{equation}

For the localization as regression task, our objective is to minimize the root mean square error between the predicted $\hat{y_{i}}$ and the reference $y_{i}$ location:
\begin{equation}
\mathcal{L}_{\textit{regression}} = \sqrt{\frac{\sum_{i} (\hat{y_{i}} - y_{i})^{2}}{2}}
\end{equation}

\subsection{Multi-task learning}

Even though the segmentation modules seem similar in form, structure and training cost, in reality, they mean two completely different things. The semantic segmentation sub-net (Stage 1), has to learn to segment the given image into pixels of different classes (e.g. road vs non-road). The second localization branch, which has the same segmentation-like cost as the first stage, is of completely different nature. Its task is to produce a map that represents the whole region. It outputs a segmentation, where white pixels mean probable locations of the input image. Our approach to localization as segmentation is quite unique and, as it turns out, it produces excellent results in practice.

For the localization task as regression, the approach is more traditional, with the x-y values (between 0 and 100 for each) produced by the regression branch meaning the relative locations, within the very large 100x100 $km^2$ wide city region.

\vspace{0.15cm}
\noindent\textbf{Optimization.} In order to train our models we use the Keras \cite{chollet2015keras} deep learning framework with the Tensorflow backend. For the first stage of our network, we start with a learning rate of 1e-4 and decrease it, no more than five times when optimization reaches a plateau. For both branches in the second stage, we do not decay the learning rate.
Training is done using the early stopping paradigm. We monitor the error on the validation set and suspend the training when the loss has not decayed for 10 epochs. We chose RMSprop\cite{hinton2012neural} as our optimizer for all stages. 
%% Emil Note: Do we have some data here? Or are we already out of space for this paper?

\section{Experimental analysis}

In our experiments, we test the capabilities of our MSMT network for both segmentation as well as geolocalization. For the task of semantic segmentation in aerial images, we tested our method on three publicly available datasets. For the geolocalization task we collected our own images from randomly sampled locations covering a city-wide area. The particularities of each dataset, as well as the training and validation setups are detailed below.

\vspace{0.15cm}
\noindent\textbf{Massachusetts Buildings Dataset.} Published in 2013, this set is among the first publicly available aerial image dataset used for training CNNs~\cite{mnih2013machine}. It consists of 151 aerial images of 1500 x 1500 pixels$^2$, at a spatial resolution of 1 m$^2$ per pixel, covering about 340 km$^2$ from the City of Boston. It offers pixelwise ground truth for two semantic classes: buildings and non-buildings. The set is randomly divided by the authors in 137 images used for training, 4 images used for validation and 10 images used for testing.

\vspace{0.15cm}
\noindent\textbf{Inria Aerial Image Labeling Dataset for Buildings.} The dataset was recently released~\cite{maggiori2017can} and is specially designed for advancing technologies in automatic pixelwise labeling of aerial imagery. It consists of 360 aerial orthorectified RGB images of 5000 x 5000 pixels$^2$, covering a surface of 1500 x 1500 m$^2$ with a spatial resolution of 30 cm per pixel. The total area coverage is 810 km$^2$, equally divided for training and testing. Each aerial image has an accompanying identically sized binary label which indicates whether a pixel belongs either to the \textit{building} or \textit{not building} class. The ground truth data is publicly disclosed only for the training subset of 180 images. The dataset covers dissimilar urban landscapes, ranging from highly dense metropolitan financial districts to alpine resorts. Regions such as Austin, Chicago, Kitsap Country, Western Tyrol and Vienna, each with 36 tiles. Following the authors' suggestion, we have selected the first five images of every location from the training set and used them for validation. 

\vspace{0.15cm}
\noindent\textbf{European Roads Dataset.} This image set was specially designed for road segmentation using only aerial RGB images. The labels were retrieved using centerline vectors from OpenStreetMap (OSM) \cite{OpenStreetMap}, having a width of approximately 5 meters on the raster image. Originally, the dataset offered only road footprints \cite{AAAIW1715177}. It was further extended by \cite{costea2017creating} for intersection detection, with regenerated, wider road labels (26 meters), in order to add contextual information, according to the authors. We have used the second setup in our experiments. The dataset contains a total of 270 images of size 1550x1600 pixels$^2$, 200 of which were used for training, 20 for validation and we reported our results on the 50 test images. Images have spatial resolution of 1 m$^2$ per pixel and cover around 374 km$^2$. We extracted 512x512 minimum overlapping tiles to train our network.

\begin{figure*}[h!]
\centering
\includegraphics[scale=0.33]{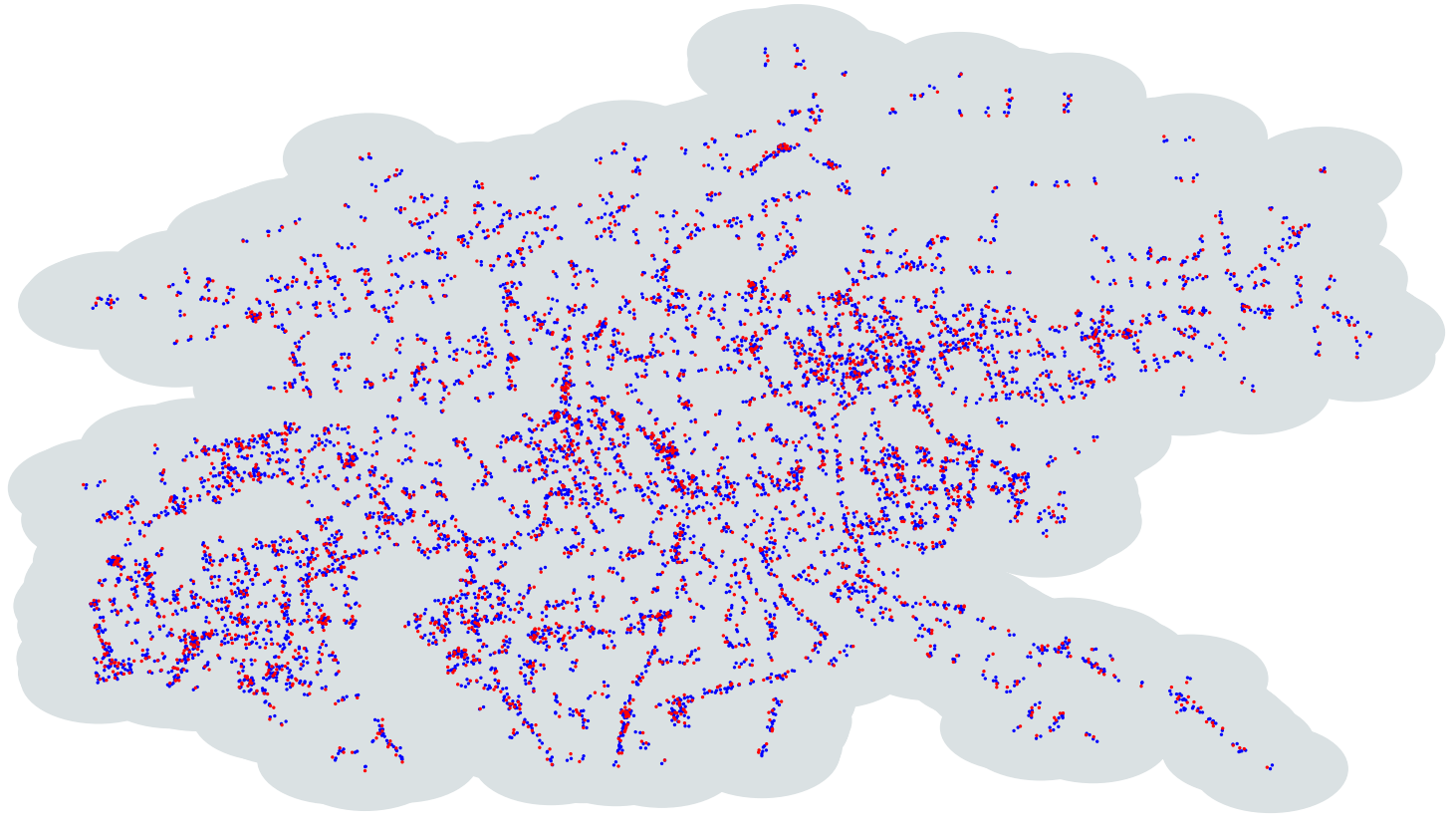}
\caption{\label{fig:data_distribution} Data distribution for our Aerial Image Localization Dataset. Training image centers are represented with blue and test image centers with red. Each grey disk depicts a region of 500 meters radius around the training and testing data -- an approximate figure for the coverage of an image. The whole imagery covers around 70km $^2$.}
\end{figure*}

\vspace{0.15cm}
\noindent\textbf{Aerial Image Localization Dataset.} Unfortunately, localization from aerial images lacks publicly available datasets. We first considered the dataset published \cite{costea2016aerial} for vision-based localization. However, images in that set are centered on intersections, which makes the task easier and not so realistic for practical applications. Therefore we decided to create our own dataset.  
In order to advance further research in this area, we will make our dataset public.
We collected 9531 512 x 512 pixels$^2$ images randomly chosen within a 100x100 m$^2$ square area around any intersection, and covering in total an European urban area of around 70 km$^2$. For training, we used 90$\%$ of the images, and for testing we used 10$\%$, randomly split. The spatial resolution of these images is 1 m$^2$ per pixel. The road footprints from OpenStreetMap were rendered with a thickness of 5 meters and used as labels. The area covered by our localization dataset is shown in Figure \ref{fig:data_distribution}.

\subsection{Semantic segmentation in aerial images}

In remote sensing, the precision/recall curve is the most frequently used evaluation metric for detection performance \cite{kn:saito2015building}. Given the difficulty of correctly labeling pixels in aerial images, Mnih et al. \cite{mnih2013machine} introduced a relaxation factor. The relaxed measure does not penalize the model for mistakes made within a range of $\rho$ pixels from its predictions.
To assess the performance of our model and enable fair comparisons with existing methods, we use the relaxed F1-score. We report the results with zero relaxation ($\rho = 0$) and also with a relaxation factor of 3 pixels ($\rho = 3$), which is consistent with previous works (\cite{kn:mnih2010learning}, \cite{kn:saito2015building}, \cite{hamaguchi2017effective}).

\setlength{\tabcolsep}{4pt}
\begin{table}
\begin{center}
\caption{Quantitative results on the Massachusetts Buildings Dataset. Higher is better.}
\label{table:mass_results}
\begin{tabular}{ll>{\centering\arraybackslash}m{2cm}>{\centering\arraybackslash}m{2cm}}
\hline\noalign{\smallskip}
Dataset & Method & \multicolumn{2}{c}{F-measure} \\
%\noalign{\smallskip} \\
& & {Relaxed 0} & {Relaxed 3} \\
\hline
\noalign{\smallskip}
Buildings & Deeplab \cite{chen2016deeplab} & - & 89.7 \% \\
& Mnih et al. \cite{mnih2013machine} & - & 91.5 \% \\
%& Sherrah \cite{sherrah2016fully} & - & 93.0 \% \\
%& FCN-8s \cite{long2015fully} & -  & 93.1 \% \\
& U-Net \cite{ronneberger2015u} & - & 94.1 \% \\
& Saito et al. \cite{saito2016multiple} & - & 94.3 \% \\
& Marcu et al. \cite{AAAIW1715177} & - & 94.3 \% \\
& Hamaguchi et al. \cite{hamaguchi2017effective} & - & 94.3 \% \\
\hline
& MTMS-Stage-1 (Ours) & 83.39 \% & \textbf{96.04} \% \\
\hline
\end{tabular}
\end{center}
\vspace*{-10mm}%Put here to reduce too much white space after table 

\end{table}
\setlength{\tabcolsep}{1.4pt}

\noindent\textbf{Results.}
We set a new state-of-the art result on the Massachusetts Buildings dataset, outperforming others by a significant margin, as shown in Table \ref{table:mass_results}. Our model uses a single RGB image and class label. We do not train a network ensemble or use multiple classes \cite{saito2016multiple}.   

\begin{figure*}[h!]
\centering
\includegraphics[scale=0.41]{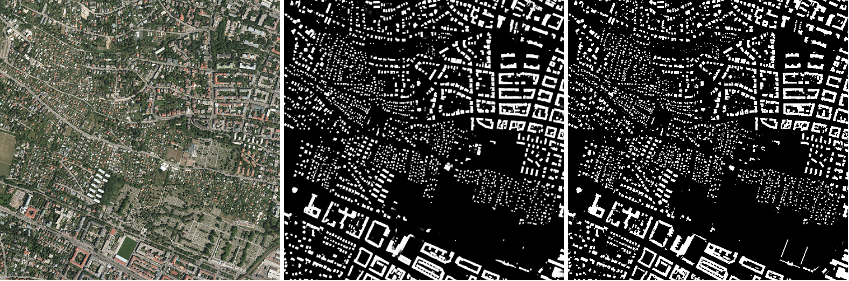}
\caption{\label{fig:inria} Qualitative results of buildings segmentations on the Inria Aerial Image Labeling Dataset. From left to right, in order, we present the 5000x5000 RGB input image, the prediction of our MSMT-Stage-1 and ground truth. Note the significant scale difference and the variety of structure between instances.}
\end{figure*}

We report \textit{IoU} performance for the Inria dataset, computed as the number of pixels labeled as building in both the prediction and the reference, divided by the number of pixels in the prediction or the reference and the \textit{Accuracy} as the percentage of correctly classified pixels. We achieve state-of-the-art results on the validation dataset using our MSMT-Stage-1 network.

\setlength{\tabcolsep}{0.5pt}
\begin{table}
\begin{center}
\caption{Segmentation results on Inria Aerial Image Labeling Dataset. Results reported on the validation set. Higher is better.}
\label{table:inria_segm_results}
\begin{tabular}{>{\centering\arraybackslash}m{3cm}>{\centering\arraybackslash}m{1.3cm}>{\centering\arraybackslash}m{1.3cm}>{\centering\arraybackslash}m{1.3cm}>{\centering\arraybackslash}m{1.3cm}>{\centering\arraybackslash}m{1.3cm}>{\centering\arraybackslash}m{1.3cm}>{\centering\arraybackslash}m{1.3cm}}
\hline\noalign{\smallskip}
Method & & Austin & Chicago & Kitsap Co. & West Tyrol & Vienna & Overall \\
\noalign{\smallskip}
\hline
\noalign{\smallskip}
\multirow{2}{*}{MLP \cite{maggiori2017can}} & IoU & 61.20 & 61.30 & 51.50 & 57.95 & 72.13 & 64.67 \\
    & Acc. & 94.20 & 90.43 & 98.92 & 96.66 & 91.87 & 94.42 \\
\hline
\multirow{2}{*}{Mask R-CNN \cite{he2017mask}} & IoU & 65.63 & 48.07 & 54.38 & 70.84 & 64.40 & 59.53 \\
    & Acc. & 94.09 & 85.56 & 97.32 & 98.14 & 87.40 & 92.49 \\
\hline
\multirow{2}{*}{SegNet MT-Loss \cite{bischke2017multi}} & IoU & \textbf{76.76} & 67.06 & \textbf{73.30} & 66.91 & 76.68 & 73.00  \\
    & Acc. & 93.21 & \textbf{99.25} & 97.84 & 91.71 & \textbf{96.61} & 95.73 \\
\hline\hline
\multirow{2}{*}{MSMT-Stage-1} & IoU & 75.39 & \textbf{67.93} & 66.35 & \textbf{74.07} & \textbf{77.12} & \textbf{73.31} \\
    & Acc. & \textbf{95.99} & 92.02 & \textbf{99.24} & \textbf{97.78} & 92.49 & \textbf{96.06} \\
\hline
\end{tabular}
\end{center}
\vspace*{-15mm}%Put here to reduce too much white space after table 
\end{table}
\setlength{\tabcolsep}{0.5pt}

\setlength{\tabcolsep}{4pt}
\begin{table}
\begin{center}
\caption{Quantitative results for the European Roads Dataset. For comparison, the previous state-of-the-art results were provided by the original paper \cite{costea2017creating} with relax 3. Higher is better.}
\label{table:eu_roads_table}
\centering\begin{tabular}{l|>{\centering\arraybackslash}m{4cm}}
\hline\noalign{\smallskip}
Method & F-measure (Relaxed 3) \\
\noalign{\smallskip}
\hline
\noalign{\smallskip}
pix2pix \cite{isola2016image} & 77.70 \%\\
U-Net \cite{ronneberger2015u} & 79.79 \%\\
LG-Seg-ResNet-IL \cite{AAAIW1715177} & 81.06 \%\\
DH-GAN \cite{costea2017creating} & \textbf{84.05} \%\\
\hline
MTMS-Stage-1 (Ours) & 82.22 \% \\
\hline
\end{tabular}
\end{center}
\vspace*{-7mm}%Put here to reduce too much white space after table 
\end{table}
\setlength{\tabcolsep}{1.4pt}

In Tables \ref{table:eu_roads_table} and \ref{table:cluj_roads_table}, we show additional results on semantic segmentation of roads. While on the European Roads Dataset our method is close to state-of-the-art, on our localization dataset we achieve near perfect F-measure (99.88$\%$), when a 3-pixel relaxation is used. 

\setlength{\tabcolsep}{4pt}
\begin{table}
\begin{center}
\caption{Roads segmentation results our Aerial Image Localization Dataset, compared with other well-known segmentation frameworks. Higher is better.}
\label{table:cluj_roads_table}
\begin{tabular}{l|>{\centering\arraybackslash}m{1.5cm}>{\centering\arraybackslash}m{1.5cm}}
\hline\noalign{\smallskip}
Method & \multicolumn{2}{c}{F-measure} \\
%\noalign{\smallskip} \\
& {Relaxed 0} & {Relaxed 3} \\
\hline
\noalign{\smallskip}
pix2pix \cite{isola2016image}  & 67.25 \% & 95.90 \%\\
U-Net \cite{ronneberger2015u} & 78.17 \% & 97.93 \%\\
\hline
MSMT-Stage-1 (Ours) & \textbf{87.86} \% & \textbf{99.88} \%\\
\hline
\end{tabular}
\end{center}
\vspace*{-10mm}%Put here to reduce too much white space after table 
\end{table}
\setlength{\tabcolsep}{1.4pt}

\subsection{Localization in aerial images}

\noindent\textbf{Localization as a segmentation task.} Similar to marking a spot on a map with an X, the localization as a segmentation problem should result in a dot marked on a normalized map. The centroid of the dot represents the location on the map. The mapping from pixels in this localization map and meters on Earth was done using the WGS84 latitude-longitude system, assuming the pixels are square and the local surface is flat - an approximation that is reasonable for the 70 km$^2$ region considered. We represent the localization map of the region with a 128x128 pixels image -- that is, a pixel was approximately 80 meters wide. This architecture, termed LocDecoder-S-128, is depicted in the lower image of Stage 2 from Figure \ref{fig:msmt_v1}.

Convergence was reached at around 1000 epochs, compared to around 50 epochs required for training the semantic segmentation networks. We also attempted using higher resolutions for the output map (256x256 pixels, 512x512 pixels). Unfortunately, the convergence speed was unsatisfactory -- probably due to the larger search space.

While localization of image elements with dots has been proposed before \cite{newell2017pixels}, we extend it in the case of predicting a location over a large area of interest. The final dot selected as the predicted location was chosen to be the centroid of the largest connected component in the predicted localization segmentation map. In the rare cases when the map predicted was left blank by the neural network, the regression output was used instead. Computing final locations as a center of a large region could produce sub-pixel accuracy and improve performance. We only use one-shot, single image matching -- no surrounding imagery or any other information apart from the RGB input is needed.

\vspace{0.15cm}
\noindent\textbf{Localization with regression.} For this task, we used the same WGS84 latitude-longitude system with minimum and maximum latitude and longitude coordinates for the interest region linearly mapped to x-y coordinates between 0 to 100. While we use the same UniEncoder, we add a location decoder (LocDecoder-R-2 in Figure \ref{fig:msmt_v1}) that directly outputs the latitude and longitude. Given the difficulty of the task, we also trained this network for about 1000 epochs.

\vspace{0.15cm}
\noindent\textbf{Localization with nearest neighbor matching.} In these experiments, we compared our results with a nearest neighbor technique based on descriptors computed from higher level features from deep road segmentation networks, similar to the approach from~\cite{costea2016aerial} and originally proposed in~\cite{lin2015deep}.

For the descriptor-based nearest neighbor localization, we extract 512 real values descriptor vectors from the end of each networks' encoder for the test images and match them against the ones generated by the training images. We test with descriptors from two different architectures, namely GAN (pix2pix)~\cite{isola2016image} and our MSMT Stage-1, in order to assess the importance of two types of high level features for the task of direct nearest neighbor localization. 

\begin{figure*}[t!]
\centering
\includegraphics[scale=0.22,keepaspectratio]{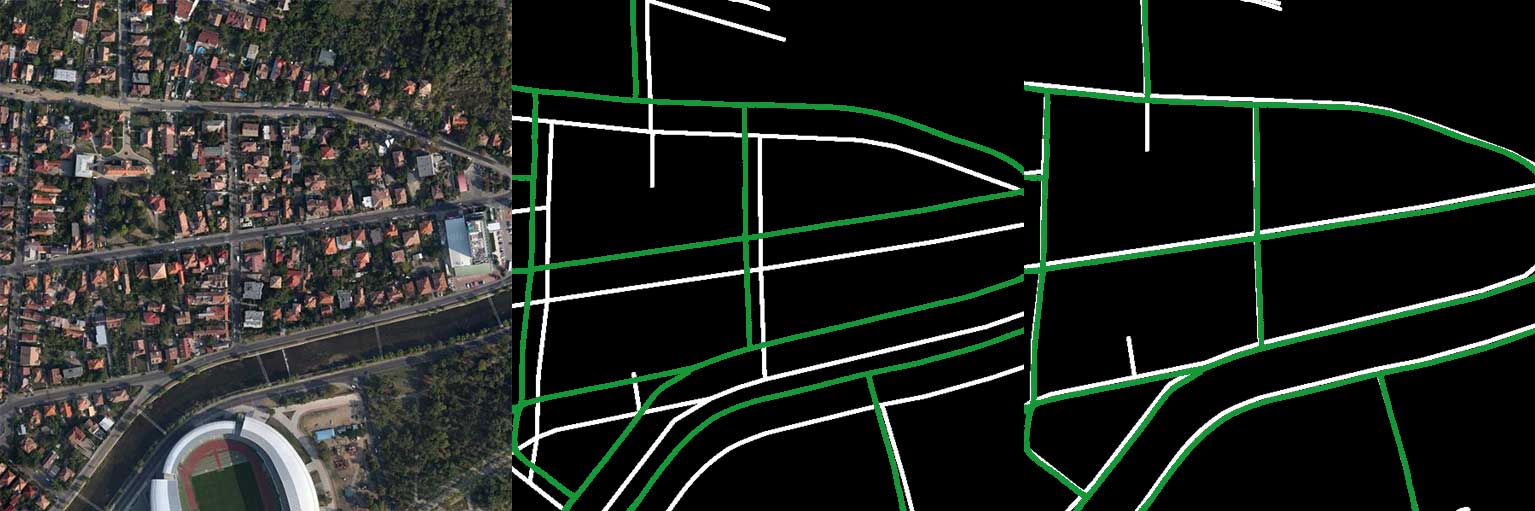}
\caption{\label{fig:pred_al_copy}Qualitative results for road detection, localization and alignment with roads from OSM. From left to right, RGB image, OSM roads (white) on top of predicted roads (green) at the location given by MSMT with LocDecoder-S-128 before alignment and OSM roads on top of predicted roads after alignment. The original error was among the highest using the segmentation method (40.38m), down to 0.32m after alignment.}
\end{figure*}

\begin{figure*}[t!]
\centering
\includegraphics[scale=0.34]{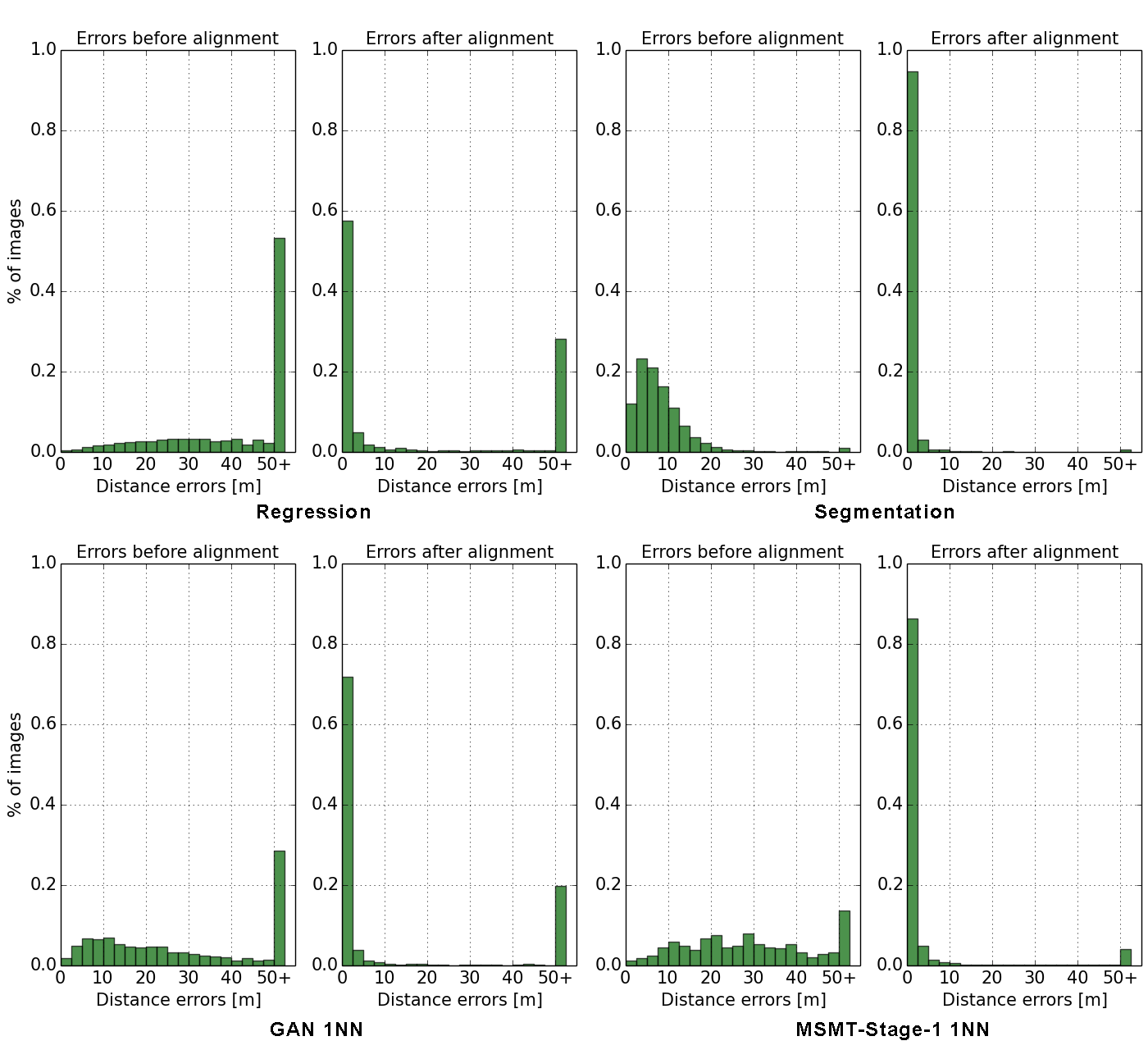}
\caption{\label{fig:reg_seg_hist_errors}Error histograms, before and after ICP alignment.
Each bin has 2.5 meters.}
\vspace*{-5mm}%Put here to reduce too much white space after table
\end{figure*}

\vspace{0.15cm}
\noindent\textbf{Alignment refinement step.} At the final step, each approach compared was then fed into a simple refinement procedure using the Iterative Closest Point (ICP) algorithm~\cite{BeMK92}, to align the road segmentation with the OpenStreetMap roads from the predicted location. As our results show (Table \ref{table:localization}, Figures \ref{fig:reg_seg_hist_errors} and \ref{fig:pred_al_copy}), the refinement step helps all methods.

In our case, we developed a simplified version of ICP that only estimates a translation between images, since they are aligned with the cardinal points, as we assume we have access to an accurate compass in the real world. The camera is also assumed to be perpendicular to the ground plane. In a real-life, UAV-use scenario, this would be accomplished using the on-board barometer to estimate the relative height and IMU to project the image onto the ground plane. 

Being a translation-only problem, the ICP method can be shown to result in a simplified solution for the following case. At each iteration, points are sampled on roads from the segmentation map produced and then on roads from the OpenStreetMap at the location predicted by the previous stages. In our experiments, points are sampled at equal distances, namely at 10 pixels. The next step is to find, for each point in the first set, its nearest neighbor in the second set. Then, we want to find the translation that minimizes the mean squared distances between the matched points. This boils down to computing the centroids of the two point sets and find the translation that takes one centroid to the other. The procedure of finding the nearest neighbors and then translate the point sets such that the centroids coincide is repeated until convergence, when the centroids are the same. Qualitative results before and after alignment are presented in Figure \ref{fig:pred_al_copy}. Beyond map resolution, further performance gains could be obtained by improving the road vector maps from OSM, that often fail to have a sub-meter alignment accuracy for road vertices.
Note that in practice, small compass and altitude errors could be corrected by first introducing small scaling and rotational noise when training the MSMT net and second, at the refinement stage, by estimating the similarity transformation and translation with ICP.

\iffalse
\setlength{\tabcolsep}{4pt}
\begin{table}
\begin{center}
\caption{Localization errors before and after alignment. Errors are expressed in meters. For MSMT with LocCombined, we combined the two localization branches as follows: while the segmentation net is on average much more precise, the regression head is used only if the segmentation produces a completely blank map. Note that for MSMT with LocDecoder-S-128 averages are computed only for the 92.7$\%$ of cases when it does not leave the out localization map blank.}
\label{table:localization}
\begin{tabular}{l>{\centering\arraybackslash}m{1.5cm}>{\centering\arraybackslash}m{1.5cm}>{\centering\arraybackslash}m{1.5cm}>{\centering\arraybackslash}m{1.5cm}}
\hline\noalign{\smallskip}
Method & \multicolumn{2}{c}{Before} & \multicolumn{2}{c}{After} \\
& {Mean} & {Median} & {Mean} & {Median} \\
%\noalign{\smallskip}
\hline
%\noalign{\smallskip}
1NN pix2pix\cite{isola2016image} & 470.50 & 53.90 & 57.97 & 1.60 \\
1NN MSMT Stage 1 & 34.11 & 31.65 & 454.79 & 1.40 \\
MSMT with LocDecoder-R-2 & 88.92 & 53.90 & 57.97 & 1.60 \\
\hline
MSMT with LocDecoder-S-128 & \textbf{9.03} & \textbf{6.85} & \textbf{1.89} & \textbf{0.75} \\
\hline
MSMT with LocCombined & 26.93 & 7.27 & 18.42 & 0.78 \\
\hline
\end{tabular}
\end{center}
\vspace*{-10mm}%Put here to reduce too much white space after table
\end{table}
\fi

\vspace{0.15cm}
\noindent\textbf{Results.}
Even though the alignment is reliable for distances smaller than 100 meters (see Figure \ref{fig:pred_al_copy} for a qualitative example of localization with and without alignment), the regression method has many outliers too far to be matched. On the other hand, using our segmentation method, 96.84$\%$ of test locations have an error of less than 20m without alignment. After alignment, 94.56$\%$ of the test locations are within 2.5m of the ground truth location and 97.58$\%$ are within 5 meters, which matches an approximate average figure for a commercial GPS \cite{Diggelen2015gpsmooc}. Regarding the 1-Nearest Neighbor error histograms for GAN and our MSMT Stage-1, only 71.69$\%$ of the GAN descriptors had an error smaller than 2.5 meters. Although the 1-NN approach using descriptors from our MSMT network perform better, (86.29$\%$ $<$ 2.5 meters), our segmentation approach has a three fold improvement in error, as shown in Figure \ref{fig:reg_seg_hist_errors}.

Although the localization as segmentation method yielded very good results, some output images were empty. From a total of 3177 test images, 2945 had a valid dot, meaning that 7.3\% of the test images were empty. In order to evaluate the method on the whole dataset, we combined the regression approach with the segmentation for the missing images, as shown in Table \ref{table:localization}.

\setlength{\tabcolsep}{4pt}
\begin{table}
\begin{center}
\caption{Inference time on GTX 1080Ti and Jetson TX2. Timings are measured in milliseconds.}
\label{table:headings}
\begin{tabular}{l>{\centering\arraybackslash}m{2cm}>{\centering\arraybackslash}m{2cm}>{\centering\arraybackslash}m{2cm}}
\hline\noalign{\smallskip}
Method & GTX 1080Ti runtime (ms) & Jetson TX2 runtime (ms) & Number of parameters \\
\noalign{\smallskip}
\hline
\noalign{\smallskip}
pix2pix \cite{isola2016image} & 18 & 280 & 23M \\
U-net \cite{ronneberger2015u} & 24 & 443 & 35M \\
MSMT-Stage-1 & 43 & 793 & 18M \\
MSMT-Stage-1-and-2 & 61 & 943 & 36M \\
\hline
\end{tabular}
\end{center}
\vspace*{-10mm}%Put here to reduce too much white space after table 
\end{table}
\setlength{\tabcolsep}{1.4pt}

%\noindent\textbf{Computation time.}
As most solutions focus on raw accuracy, we are also interested in implementing our MTMS network on a device with reduced performance, such as an embedded GPU. We therefore tested the runtime of the networks on an NVIDIA GTX 1080Ti and a Jetson TX2 architecture, which is more appropriate for a UAV use case. The results are given in Table \ref{table:headings}. Note that on Jetson TX2 timings are still under a second per image.

%% Emil-Note: here you need to make a reference to Table 7. I've added a phrase to this effect.

\section{Additional results}

\subsection{Building detection}

We present additional qualitative results of our MSMT Stage-1 network for the task of building detection on the two datasets we have experimented with, namely the Massachusetts Buildings Dataset \cite{mnih2013machine} and the Inria Aerial Image Labeling Dataset \cite{maggiori2017can}.

\begin{figure*}[h!]
\centering
\includegraphics[scale=0.19]{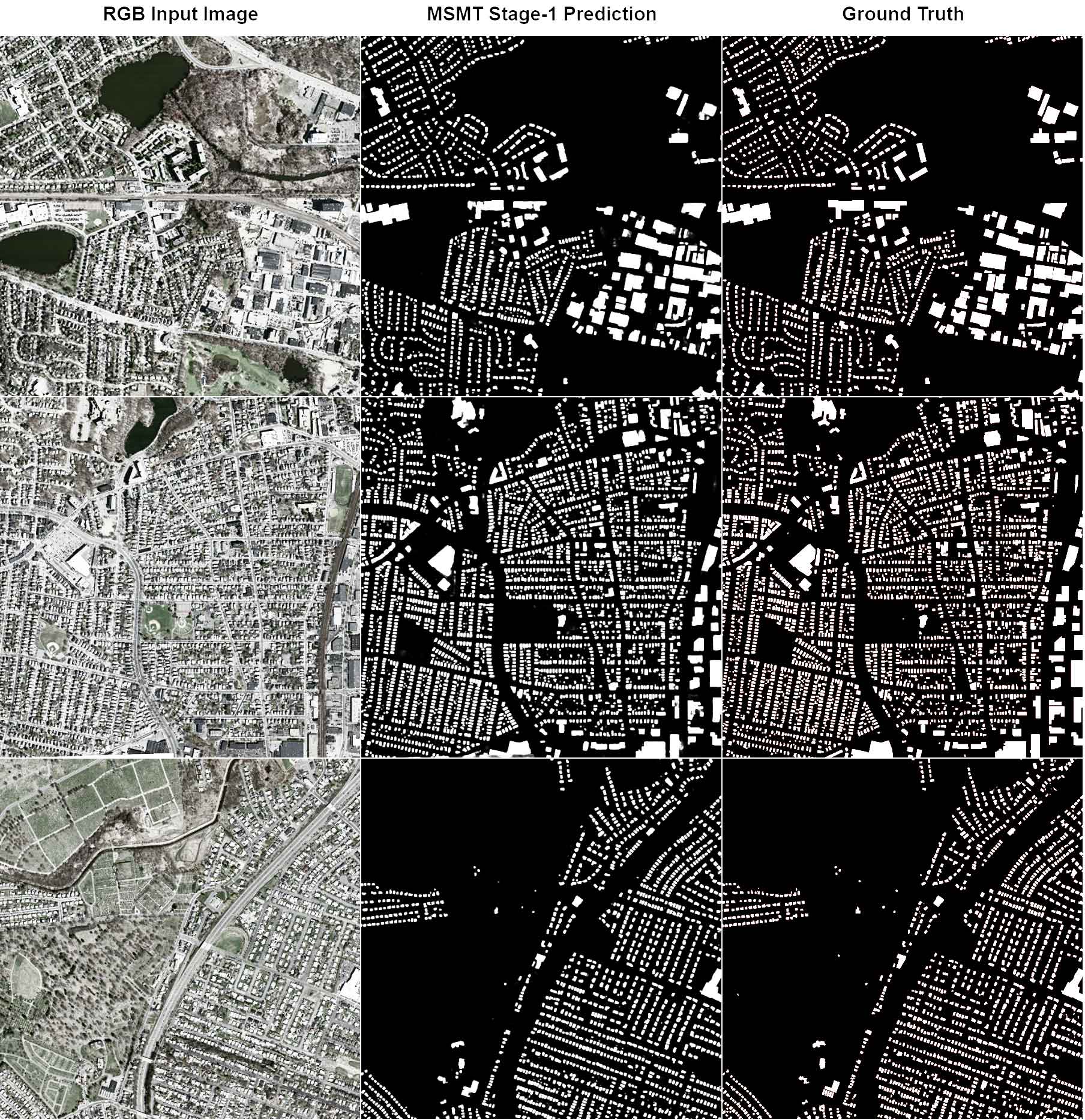}
\caption{\label{fig:mass_dataset_results} Segmentation results on the Massachusetts Buildings Dataset. From left to right, in order, we present the RGB test input image of size 1500 x 1500 pixels, followed by the soft segmentation map produced by MSMT Stage-1 and the corresponding ground truth map. Note that the two segmentations are almost identical, a fact that is reflected by our $96\%$ F1-score on this dataset.
}
\end{figure*}

\noindent\textbf{Massachusetts Buildings Dataset.} This dataset features urban and suburban images from the Boston area. Structures tagged as buildings cover a wide span of sizes, from garages to shopping malls. The labeling omission noise is around 5$\%$ \cite{mnih2013machine}. In the paper submission, we report state-of-the-art results on the test set. Qualitative results are shown in Figure \ref{fig:mass_dataset_results}.

\begin{figure*}[h!]
\centering
\includegraphics[scale=0.14]{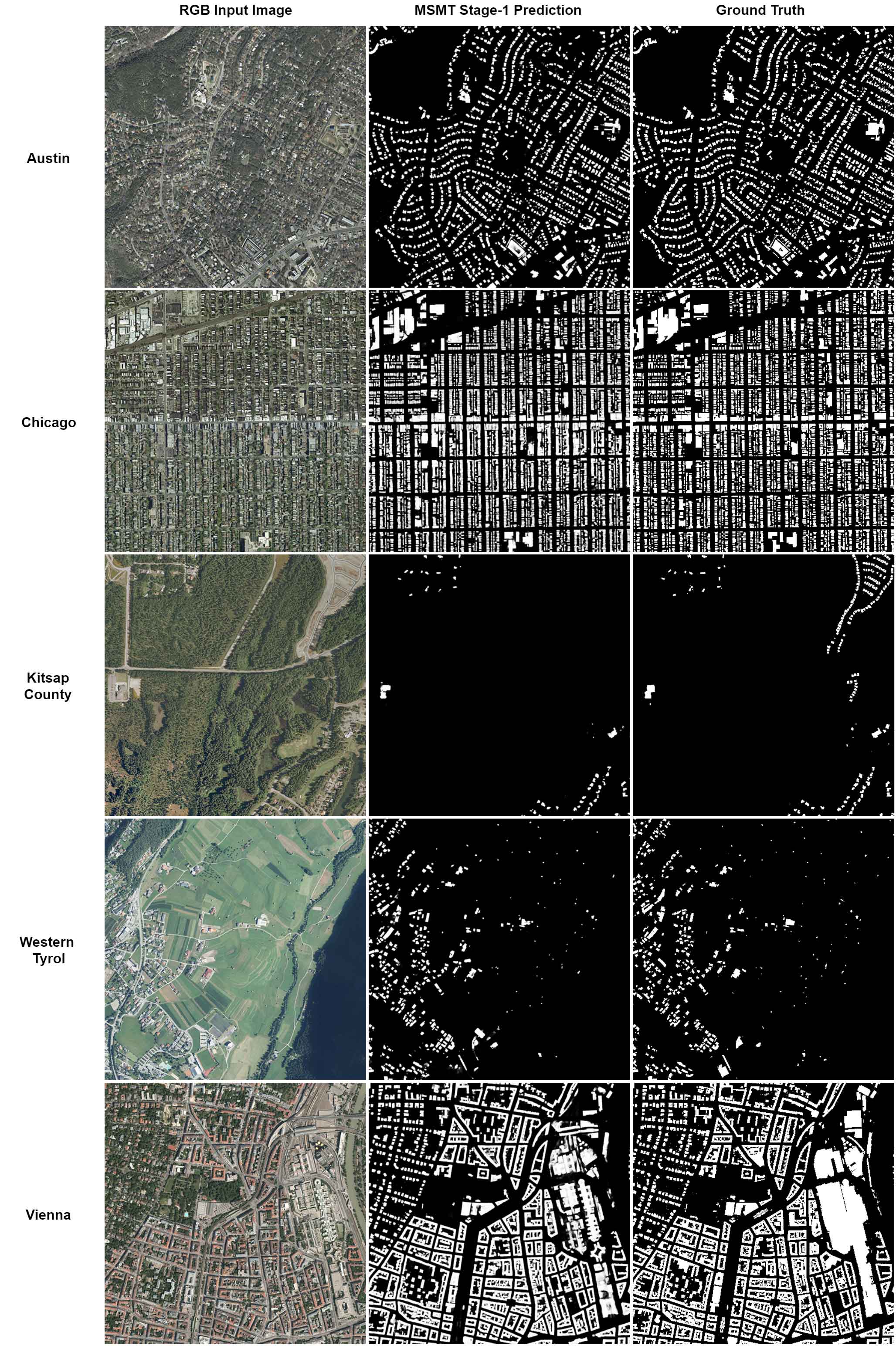}
\caption{\label{fig:inria_dataset_results} Building detection results on the recent Inria Aerial Image Labeling Dataset. The figure depicts one example from each of the 5 regions in the validation set, on which we report state-of-the-art results.}
\end{figure*}

\noindent\textbf{Inria Aerial Image Labeling.} The Inria Aerial Image Labeling dataset is diverse and difficult, with imagery from mountain resorts to dense urban areas. It features very large variations in scale (see Figure \ref{fig:inria_dataset_results}, Western Tyrol and Vienna examples) and covers dissimilar urban settlements, ranging from densely populated areas (see Figure \ref{fig:inria_dataset_results}, Chicago, Austin and Vienna examples) compared to alpine towns, such as Kitsap County and Western Tyrol.

Beside the difficulties of segmenting such varied regions, the reader can also observe ground truth label inconsistencies, which can hinder the network training and also have an impact on the evaluation score. Such an example is shown in Figure \ref{fig:inria_dataset_results}, in which an image from the Kitsap County has the whole upper right labeled region missing. Also, the largest single building label from the Vienna image depicts a very large instance structure that should be split into pieces. Our model is robust to omission noise and is also capable of segmenting complex building shapes. We report state-of-the-art results on this dataset.

%The first example depicts detection on the validation images for Vienna.

%\begin{figure*}[h!]
%#\centering
%\includegraphics[scale=0.11]{images/RGB_pred_vienna1_crop_1000.png}
%\caption{\label{fig:inria_crop_1000} Qualitative results of buildings segmentations on the Inria Aerial Image Labeling %Dataset. From left to right, in order, we present a 1000 x 1000 px crop of the RGB input image, the prediction of our %MSMT-Stage-1.}
%\end{figure*}

\begin{figure*}[h!]
\centering
\includegraphics[scale=0.165]{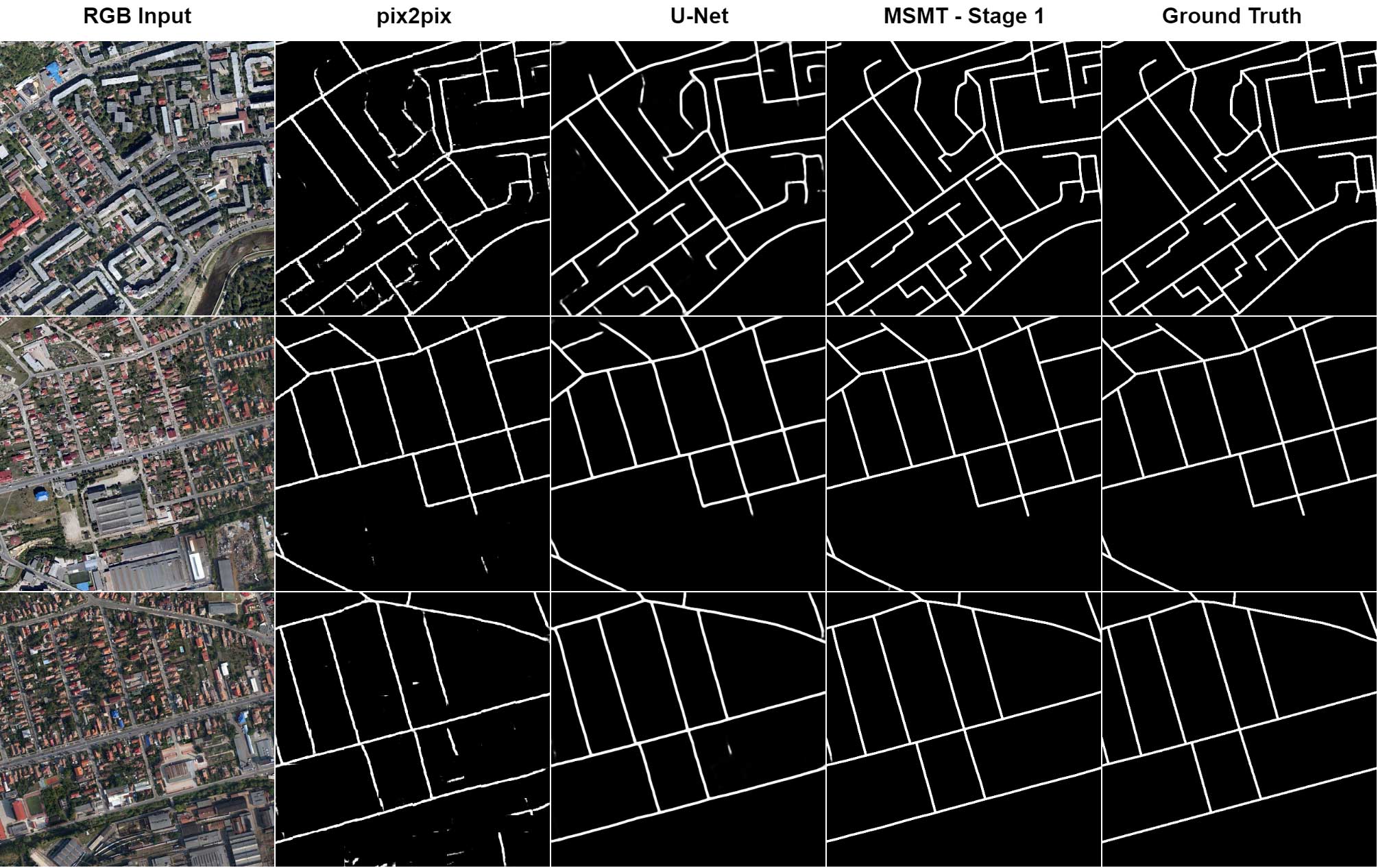}
\caption{\label{fig:localization_dataset_results} Qualitative results of road segmentations on Our Aerial Localization Dataset. Images have a size of 512 x 512 pixels. We compare our results with well-known segmentation frameworks. We obtain almost perfect results on test images, reaching 99.88\% F-measure.}
\end{figure*}

\subsection{Road detection}

We present qualitative results for road detection on two datasets, the publicly available European Roads Dataset \cite{costea2017creating} and Our Aerial Localization Dataset, which we plan to make publicly available soon\footnote{https://sites.google.com/site/aerialimageunderstanding/}.

In Figure \ref{fig:localization_dataset_results} we compare our MSMT Stage-1 semantic segmentation network with other well-known segmentation frameworks such as pix2pix \cite{isola2016image} and U-Net \cite{ronneberger2015u}, on Our Aerial Localization Dataset. The design choices for our 
MSMT Stage-1 architecture bring a notable performance improvement. It reduces the false positive rate, it has smoother road predictions and also detects complex road structures.

\begin{figure*}[h!]
\centering
\includegraphics[scale=0.25]{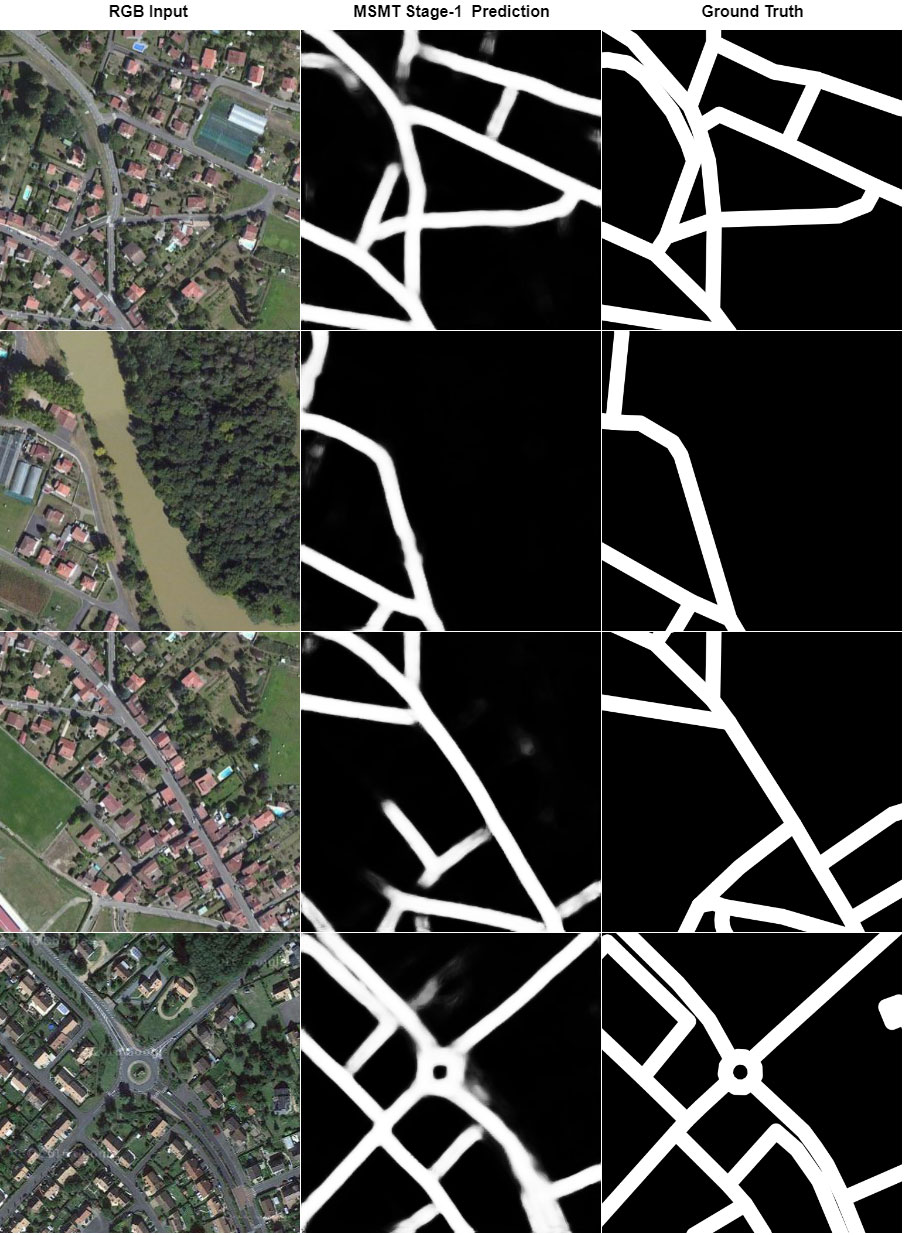}
\caption{\label{fig:eu_roads_dataset_results} Results on European Roads Dataset \cite{costea2017creating}. Our MSMT Stage-1 model has some difficulties with connecting roads (improper roads intersection detection, depicted in example 1 and 3) but it is robust to road occlusions (e.g. trees) shown in the second example.}
\end{figure*}

Results on the European Roads Dataset are shown in Figure \ref{fig:eu_roads_dataset_results}, in which we come in second place. The labels are thicker than the actual roads, since we used the same training setup as proposed by \cite{costea2017creating}. While this could prove useful for context aggregation, it leads to several failure cases and poorer overall performance for our MSMT Stage-1 architecture when compared to \cite{costea2017creating}.

\subsection{Vision-based localization}

Localization can be done in several ways within our MSMT network. One is by outputting a direct longitude and latitude location, by formulating localization as a regression task. The other strategy is to formulate localization as a segmentation problem, with an output that represents a map of the whole region (in which we localize the given individual image), with positively classified pixels representing the potential location of the image. The location in that case is estimated as the centroid of the largest connected component on that output map. As explained in the paper submission, the best results are obtained when we consider the output of the segmentation branch -- yielding higher localization accuracy -- and consider the regression one only when the segmentation fails to provide a location -- producing a blank map.

Given a predicted location, we extract a wider area from OpenStreetMap roads and align it against the predicted roads from the RGB image, in order to produce a more accurate, refined localization. For this geometric alignment, we propose a simplified ICP method, as presented in the paper.
Quantitative results show excellent performance on image alignment for up to 100 meters initial localization error, predicted by the MSTM network.

\begin{figure*}[h!]
\centering
\includegraphics[scale=0.22]{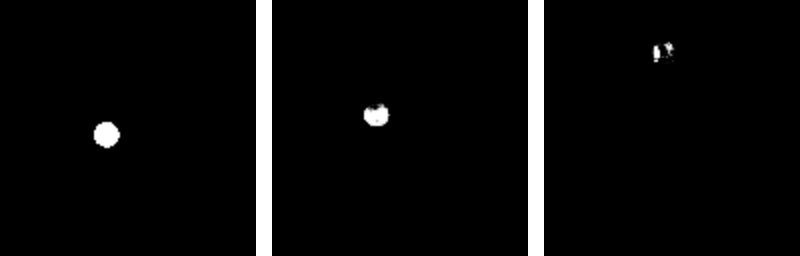}
\caption{\label{fig:dot_localization} Qualitative results of segmentation-based localization. We present different cases of localization maps.
From left to right, we show 1) a localization map that produces an almost perfect round dot (16 meters error before alignment); 2) a localization map with a slightly deformed dot (33 meters error before alignment) and 3) a localization map with a fragmented dot and wrong location (1 km error). The first two cases result in an almost perfect localization after alignment, while the third cannot be recovered by ICP.}
\end{figure*}

\begin{figure*}[h!]
\centering
\includegraphics[scale=0.15]{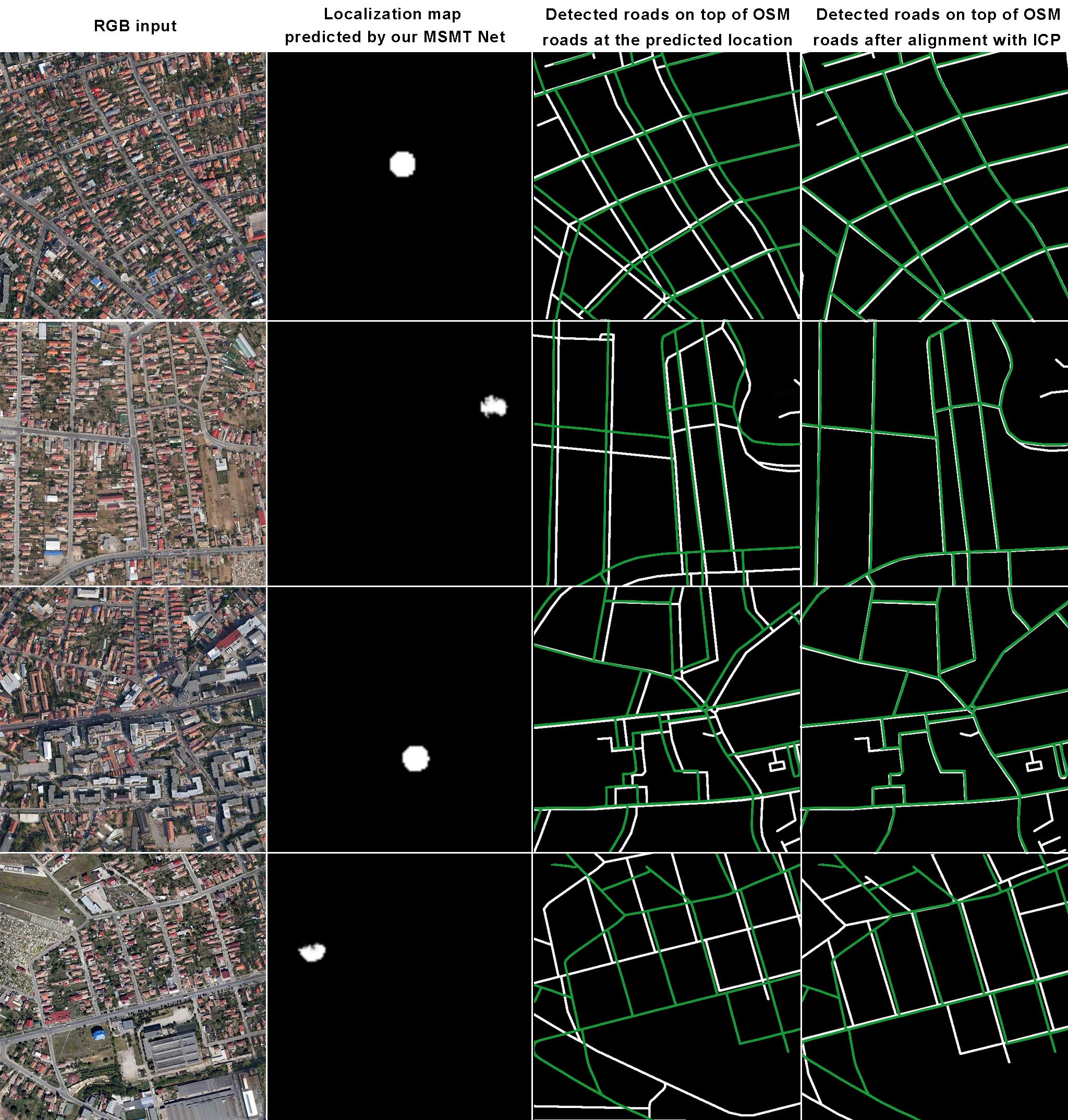}
\caption{\label{fig:alignment} Qualitative results of alignment. From left to right, we show 1) the input test RGB image; 2) location prediction as a localization map over the whole area -- with positive pixels (white dot) representing possible locations; 3) roads predicted by MSMT Stage-1 (green) on top of OSM roads (white) at the location predicted by MSMT, before the alignment; 4) final localization, after alignment with ICP, with the two road maps, the predicted and the OSM (ground truth) overlapped. The last row shows a failed case of localization, which is due to the presence of symmetries and repeated patterns. These cases also pose difficult challenges in geometric alignment tasks, in general.}
\end{figure*}

In Figure \ref{fig:alignment} we present qualitative results for localization and alignment, to better exemplify how the system works. We show three success cases, and a particular fail case, where the grid pattern affects the performance of our alignment method.

\noindent \textbf{Specific details on localization with segmentation.}
As previously shown, this type of localization is trained (from many pairs of images with locations) to place a dot centered on the actual location of the image, within a certain region -- where the region is represented as a 128 x 128 pixels map. In this case of localization with segmentation, around 6\% of test cases result in an empty label. In the cases when positively labeled pixels are produced, the location is estimated as the center of the largest connected component. The accuracy is very high in this case. Before the alignment the average is $9.3$ meters and after the alignment it is $1.89$ meters, as described in Table 6 of the paper submission. Even in difficult cases, the map being produced concentrates around a single location, as shown in Figure~\ref{fig:dot_localization}.

\section{Conclusions}

We propose a multi-stage, multi-task deep neural network that offers certain advantages over the existing literature in aerial image recognition and geolocalization. 
It has a unified and efficient structure that permits, at test time, to solve in one forward pass both problems of semantic segmentation and geolocalization.
The combined recognition-localization capability within a single and efficient structure, makes the approach potentially useful for low computational cost real-world UAV systems. Another structural advantage of the MSMT network is the efficient re-use of the UniEncoder for all three tasks of semantic segmentation, localization as regression and localization as segmentation. We prove the value of our contributions through extensive experiments -- on semantic segmentation we achieve state of the art results on two challenging datasets, the Inria Aerial Image Labeling dataset and Massachusetts Buildings. On geolocalization, we reach near commercial GPS-level accuracy, with more than 97$\%$ of cases having a localization error of less than 5 meters.

As future work, we plan to add more high-level features to the input of the localization branch (e.g., buildings). We believe that the capability to segment the image into multiple classes simultaneously (e.g. houses, large buildings and constructions, green areas, water, agricultural regions)  could be used to better learn localization if fed into the second localization step. We plan to use such multiple semantic segmentations, for which we also target a lower altitude regime from different view-points and angles that is more suitable for practical UAV applications. Our final step of accurate alignment, while simple and effective, could also be optimized, in future work, with a more efficient neural network, trained especially for geometric alignments in cases of small displacements and transformations.

%\clearpage

\subsubsection{Acknowledgements:} This work was supported in part by UEFISCDI, project PN-III-P4-ID-ERC-2016-0007 and the Romanian Ministry of European Funds, project IAVPLN POC-A1.2.1D-2015-P39-287.

\bibliographystyle{splncs03}
\bibliography{egbib}

\end{document}